\documentclass[10pt,twocolumn,letterpaper]{article}

\usepackage{cvpr}
\usepackage{times}
\usepackage{epsfig}
\usepackage{graphicx}
\usepackage{amsmath}
\usepackage{amssymb}

\newcommand{\myref}[1]{Eq.~\eqref{#1}}

\usepackage{enumerate}
\usepackage{enumitem}
\usepackage{multirow}
\usepackage{caption, subcaption}
\usepackage{booktabs}

\usepackage[pagebackref=true,breaklinks=true,letterpaper=true,colorlinks,bookmarks=false]{hyperref}

\cvprfinalcopy 

\def\cvprPaperID{5028} 

\begin{document}

\title{Learning Actor Relation Graphs for Group Activity Recognition}

\author{Jianchao Wu \quad Limin Wang \quad Li Wang \quad Jie Guo \quad Gangshan Wu\\
State Key Laboratory for Novel Software Technology, Nanjing University, China\\
}

\maketitle

\begin{abstract}
Modeling relation between actors is important for recognizing group activity in a multi-person scene.
This paper aims at learning discriminative relation between actors efficiently using deep models. To this end, we propose to build a flexible and efficient {\rm Actor Relation Graph} (ARG) to simultaneously capture the appearance and position relation between actors. Thanks to the Graph Convolutional Network, the connections in ARG could be automatically learned from group activity videos in an end-to-end manner, and the inference on ARG could be efficiently performed with standard matrix operations. Furthermore, in practice, we come up with two variants to sparsify ARG for more effective modeling in videos: spatially localized ARG and temporal randomized ARG. We perform extensive experiments on two standard group activity recognition datasets: the Volleyball dataset and the Collective Activity dataset, where state-of-the-art performance is achieved on both datasets.
We also visualize the learned actor graphs and relation features, which demonstrate that the proposed ARG is able to capture the discriminative relation information for group activity recognition.~\footnote{The code is available at \url{https://github.com/wjchaoGit/Group-Activity-Recognition}}
\end{abstract}

\section{Introduction}
\begin{figure}[t]

\setlength{\abovecaptionskip}{-0.2cm}

\begin{center}
   \includegraphics[width=1\linewidth]{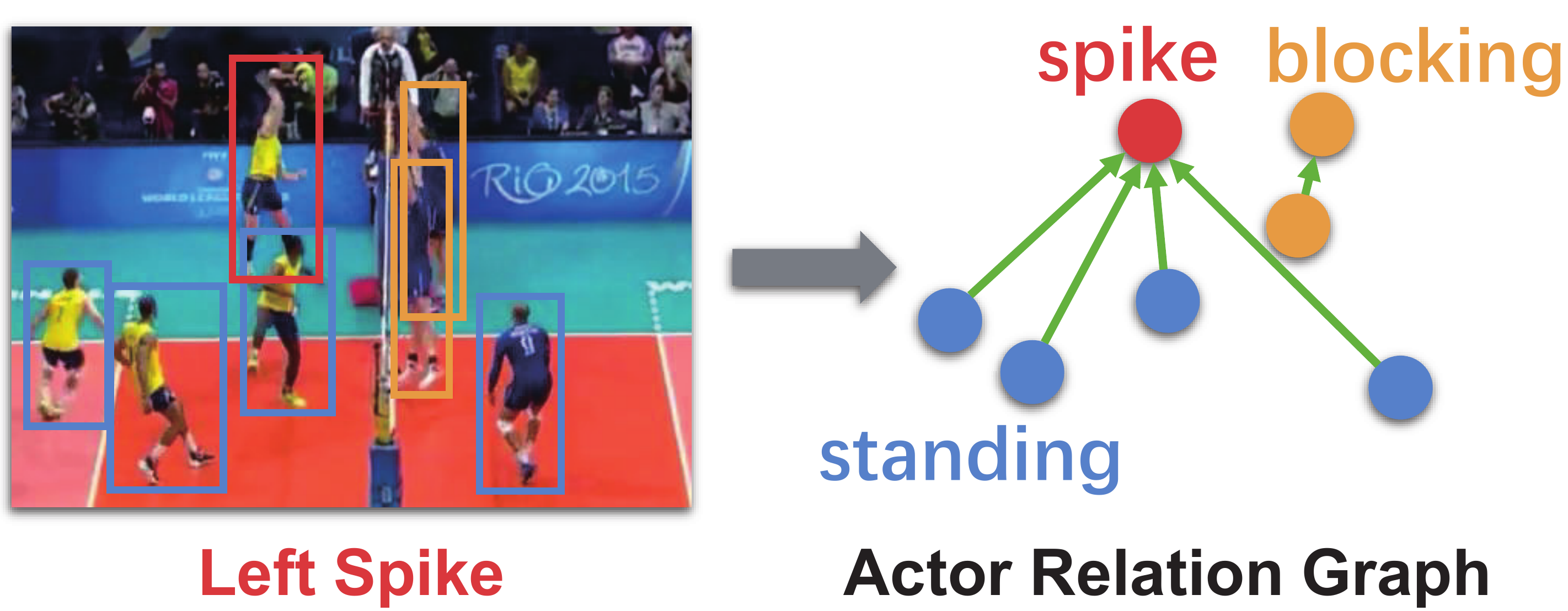}
\end{center}
   \caption{Understanding group activity in multi-person scene requires accurately determining relevant relation between actors. Our model learns to represent the scene by actor relation graph, and performs reasoning about group activity (``left spike" in the illustrated example) according to the graph structure and nodes features. Each node denotes an actor, and each edge represents the relation between two actors}
   \vspace{-4mm}
\label{fig:figure_1}
\end{figure}

Group activity recognition is an important problem in video understanding~\cite{WangL0G18,SimonyanZ14,TranBFTP15,gan2015devnet} and has many practical applications, such as surveillance, sports video analysis, and social behavior understanding. To understand the scene of multiple persons, the model needs to not only describe the individual action of each actor in the context, but also infer their collective activity. 
The ability to accurately capture relevant relation between actors and perform relational reasoning is crucial for understanding group activity of multiple people~\cite{cite_a_12, cite_a_10, cite_d_1, cite_a_17, cite_a_18, cite_a_5, cite_a_2, cite_a_7}. 
However, modeling the relation between actors is challenging, as we only have access to individual action labels and collective activity labels, without knowledge of the underlying interaction information. 
It is expected to infer relation between actors from other aspects such as {\em appearance similarity} and {\em relative location}. 
Therefore, it is required to model these two important cues when we design effective deep models for group activity understanding.

Recent deep learning methods have shown promising results for group activity recognition in videos~\cite{cite_a_1,cite_a_2,cite_a_4,cite_a_5,cite_a_6,cite_a_7,cite_a_17,cite_a_18}. Typically, these methods follow a two-stage recognition pipeline. First, the person-level features are extracted by a convolutional neural network (CNN). Then, a global module is designed to aggregate these person-level representations to yield a scene-level feature. Existing methods model the relation between these actors with an inflexible graphical model~\cite{cite_a_17}, whose structure is manually specified in advance, or using complex yet unintuitive message passing mechanism~\cite{cite_a_5,cite_a_18}. To capture temporal dynamics, a recurrent neural network (RNN) is usually used to model temporal evolution of densely sampled frames~\cite{cite_a_1,cite_a_2}. These models are generally expensive at computational cost and sometimes lack the flexibility dealing with group activity variation.

In this work, we address the problem of capturing appearance and position relation between actors for group activity recognition. Our basic aim is to model actor relation in a more flexible and efficient way, where the graphical connection between actors could be automatically learned from video data, and inference for group activity recognition could be efficiently performed.
Specifically, we propose to model the actor-actor relation by building a {\em Actor Relation Graph} (ARG), illustrated in Figure~\ref{fig:figure_1}, where the node in the graph denotes the actor's features, and the edge represents the relation between two actors. The ARG could be easily placed on top of any existing 2D CNN to form a unified group activity recognition framework. Thanks to the operation of graph convolution~\cite{cite_g_10}, the connections in ARG can be automatically optimized in an end-to-end manner. Thus, our model can discover and learn the potential relations among actors in a more flexible way. Once trained, our network can not only recognize individual actions and collective activity of a multi-person scene, but also on-the-fly generate the video-specific actor relation graph, facilitating further insights for group activity understanding.
 
To further improve the efficiency of ARG for long-range temporal modeling in videos, we come up with two techniques to sparsify the connections in ARG. Specifically, in spatial domain, we design a {\em localized} ARG by forcing the connection between actors to be only in a local neighborhood. For temporal information, we observe that slowness is naturally video prior, where frames are densely captured but semantics varies very slow. Instead of connecting any pair frame, we propose a {\em randomized} ARG by randomly dropping several frames and only keeping a few. This random dropping operation is able to not only greatly improve the modeling efficiency but also largely increase the diversity of training samples, reducing the overfitting risk of ARG.

In experiment, to fully utilize visual content, we empirically study different methods to compute pair-wise relation from the actor appearance features.
Then we introduce constructing multiple relation graphs on an actors set to enable the model to focus on more diverse relation information among actors.
We report performance on two group activity recognition benchmarks: the Volleyball dataset~\cite{cite_d_2} and the Collective Activity dataset~\cite{cite_d_1}. Our experimental results demonstrate that our ARG is able to obtain superior performance to the existing state-of-the-art approaches.

The major contribution of this paper is summarized as follows:
\begin{itemize}
\item We construct flexible and efficient actor relation graphs to simultaneously capture the appearance and position relation between actors for group activity recognition. It provides an interpretable mechanism to explicitly model the relevant relations among people in the scene, and thus the capability of discriminating different group activities.
\item We introduce an efficient inference scheme over the actor relation graphs by applying the GCN with sparse temporal sampling strategy. The proposed network is able to conduct relational reasoning over actor interactions for the purpose of group activity recognition.
\item The proposed approach achieves the state-of-the-art results on two challenging benchmarks: the Volleyball dataset~\cite{cite_d_2} and the Collective Activity dataset~\cite{cite_d_1}. Visualizations of the learned actor graphs and relation features show that our approach has the ability to attend to the relation information for group activity recognition.
\end{itemize}

\begin{figure*}[t]

\setlength{\abovecaptionskip}{-0.2cm}

\begin{center}

\includegraphics[width=1.0\linewidth]{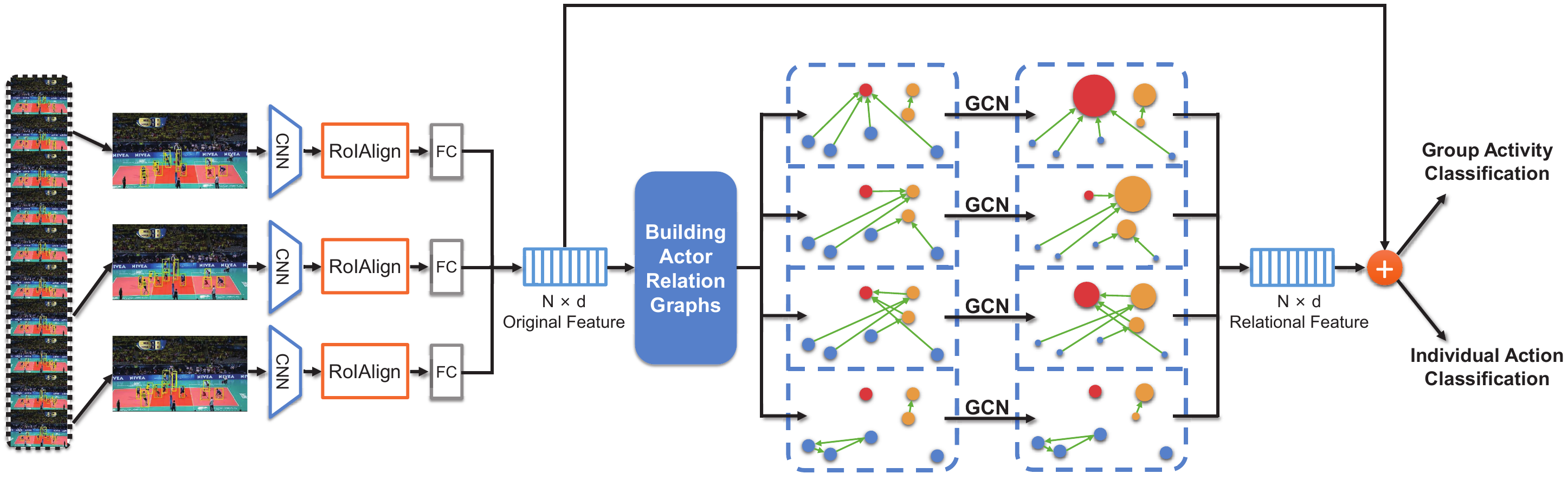}

\end{center}
\caption{An overview of our network framework for group activity recognition. 
We first extract feature vectors of actors from sampled video frames.
We use a $d$-dimension vector to represent an actor bounding box.
And the total number of bounding boxes in sampled frames equals $N$.
Multiple actor relation graphs are built to capture relation information among actors.
Afterwards, Graph Convolutional Networks are used to perform relational reasoning on graphs.
The outputs of all graphs are then fused to produce the relational feature vectors of actors.
Finally, original feature and relational feature are aggregated and fed into classifiers of group activity and individual action.
}

\label{fig:figure_2}
\end{figure*}

\section{Related Work}
{\bf Group activity recognition.}
Group activity recognition has been extensively studied from the research community.
The earlier approaches are mostly based on a combination of hand-crafted visual features with probability graphical models~\cite{cite_a_10,cite_a_11,cite_a_12,cite_a_13,cite_a_15,cite_a_16,cite_a_19}  or AND-OR grammar models~\cite{cite_a_9,cite_a_14}. 
Recently, the wide adoption of deep convolutional neural networks (CNNs) has demonstrated significant performance improvements on group activity recognition~\cite{cite_a_1,cite_a_2,cite_a_3,cite_a_4,cite_a_5,cite_a_6,cite_a_7,cite_a_17,cite_a_18}.
Ibrahim \etal ~\cite{cite_a_2} designed a two-stage deep temporal model, which builds a LSTM model to represent action dynamics of individual people and another LSTM model to aggregate person-level information.
Bagautdinov \etal ~\cite{cite_a_1} presented a unified framework for joint detection and activity recognition of multiple people.
Ibrahim \etal ~\cite{cite_a_17} proposed a hierarchical relational network that builds a relational representation for each person.
There are also efforts that explore modeling the scene context via structured recurrent neural networks ~\cite{cite_a_5,cite_a_7,cite_a_18} or generating captions ~\cite{cite_a_6}.
Our work differs from these approaches in that it explicitly models the interactions information via building flexible and interpretable ARG. Moreover, instead of using RNN for information fusion, we employ GCN with sparse temporal sampling strategy which enables relational reasoning in an efficient manner.

{\bf Visual relation.}
Modeling or learning relation between objects or entities is an important problem in computer vision~\cite{cite_r_6,cite_r_7,cite_r_8,cite_r_9,Wang0T16}.
Several recent works focus on detecting and recognizing human-object interactions (HOI)~\cite{cite_r_10,cite_r_11,cite_r_12,cite_r_13,cite_r_15}, which usually requires additional annotations of interactions.
In scene understanding, a lot of efforts have been made on modeling pair-wise relationships for scene graph generation~\cite{cite_r_1,cite_r_2,cite_r_3,cite_r_4,cite_r_5,cite_g_16}. 
Santoro \etal ~\cite{cite_r_16} proposed a relation network module for relational reasoning between objects, which achieves super-human performance in visual question answering.
Hu \etal ~\cite{cite_r_17} applied an object relation module to object detection, and verified the efficacy of modeling object relations in CNN based detection.
Besides, many works showed that modeling interactions information can help action recognition~\cite{cite_r_18,cite_r_19,cite_r_20,cite_r_21,cite_r_22}. 
We show that explicitly exploiting the relation information can achieve significant gain on group activity recognition accuracy.

{\bf Neural networks on graphs.} 
Recently, integrating graphical models with deep neural networks is an emerging topic in deep learning research.
A considerable amount of models has arisen for reasoning on graph-structured data at various tasks, such as classification of graphs
~\cite{cite_g_5,cite_g_6,cite_g_7,cite_g_8,cite_g_9},
classification of nodes in graphs
~\cite{cite_g_10,cite_g_11,cite_g_12},
and modeling multi-agent interacting physical systems
~\cite{cite_g_1,cite_g_2,cite_g_3,cite_g_13}.
In our work, we apply the Graph Convolutional Network (GCN)~\cite{cite_g_10} which was originally proposed for semi-supervised learning on the problem of classifying nodes in a graph.
There are also applications of GCNs to single-human action recognition problems~\cite{cite_g_14,cite_g_15}.
However, it would be inefficient to compute all pair-wise relation across all video-frame to build video as a fully-connected graph.
Therefore, we build multi-person scene as a sparse graph according to relative location.
Meanwhile, we propose to combine GCN with sparse temporal sampling strategy~\cite{cite_o_1} for more efficient learning.

\section{Approach}
Our goal is to recognize group activity in multi-person scene by explicitly exploiting relation information. To this end, we build {\em Actor Relation Graph} (ARG) to represent multi-person scene, and perform relational reasoning on it for group activity recognition. In this section, we will give detailed descriptions of our approach. First, we present an overview of our framework. Then, we introduce how to build ARG. Finally, we describe the efficient training and inference algorithms for ARG.

\subsection{Group Activity Recognition Framework}
\label{section:Framework}
The overall network framework is illustrated in Figure~\ref{fig:figure_2}.
Given a video sequence and the bounding boxes of the actors in the scene, our framework takes three key steps.
First, we uniformly sample a set of $K$ frames from the video and extract feature vectors of actors from sampled frames. 
We follow the feature extraction strategy used in~\cite{cite_a_1}, which adopts Inception-v3~\cite{cite_o_4} to extract a multi-scale feature map for each frame. Besides that, we also have conducted experiments on other backbone models to verify the generality and effectiveness of our approach. We apply RoIAlign~\cite{cite_o_3} to extract the features for each actor bounding box from the frame feature map. After that, a fc layer is performed on the aligned features to get a $d$ dimensional appearance feature vector for each actor. 
The total number of bounding boxes in $K$ frames is denoted as $N$.
We use a $ N \times d$ matrix $\mathbf{X}$ to represent feature vectors of actors.

Afterwards, upon these original features of actors, we build actor relation graphs, where each node denotes an actor.
Each edge in the graphs is a scalar weight, which is computed according to two actors' appearance features and their relative location.
To represent diverse relation information, we construct multiple relation graphs from a same set of actors features.

Finally, we perform learning and inference to recognize individual actions and group activity.
We apply the GCN to conduct relational reasoning based on ARG.
After graph convolution, the ARGs are fused together to generate relational representation for actors, which is also in $N \times d$ dimension. 
Then two classifiers respectively for recognizing individual actions and group activity will be applied on the pooled actors' relational representation and the original representation.
We apply a fully connected layer on individual representation for individual action classification. The actor representations are maxpooled together to generate scene-level representation, which is used for group activity classification through another fully connected layer.

\subsection{Building Actor Relation Graphs}
\label{section:Graph}

As mentioned above, ARG is the key component in our framework.
We utilize the graph structure to explicitly model pair-wise relation information for group activity understanding.
Our design is inspired by the recent success of relational reasoning and graph neural networks~\cite{cite_r_16,cite_g_10}.

{\bf Graph definition.}
Formally, the nodes in our graph correspond to a set of actors $A=\{(\mathbf{x}_i^a,\mathbf{x}_i^s)|i=1,\cdots,N\}$, where $N$ is the number of actors, $\mathbf{x}_i^a \in \mathbb{R}^{d}$ is actor $i$'s appearance feature, and and $\mathbf{x}_i^s=(t_i^x,t_i^y)$ is the center coordinates of actor $i$'s bounding box. We construct graph $\mathbf{G} \in \mathbb{R}^{N \times N}$ to represent pair-wise relation among actors, where relation value $\mathbf{G}_{ij}$ indicates the importance of actor $j$'s feature to actor $i$. 

In order to obtain sufficient representational power to capture underlying relation between two actors, both appearance features and position information need to be considered. Moreover, we note that appearance relation and position relation have different semantic attributes. To this end, we model the appearance relation and position relation in a separate and explicit way. The relation value is defined as a composite function below:
\begin{equation}
    \mathbf{G}_{ij}=h\left( f_a(\mathbf{x}_i^a,\mathbf{x}_j^a),f_s(\mathbf{x}_i^s,\mathbf{x}_j^s) \right) ,  
\label{eq:graph}
\end{equation}
where $f_a(\mathbf{x}_i^a,\mathbf{x}_j^a)$ denotes the appearance relation between two actors, and the position relation is computed by $f_s(\mathbf{x}_i^s,\mathbf{x}_j^s)$. The function $h$ fuses appearance and position relation to a scalar weight.

In our experiments, we adopt the following function to compute relation value:
\begin{equation}
    \mathbf{G}_{ij}=\frac{ f_s(\mathbf{x}_i^s,\mathbf{x}_j^s) ~ \mathrm{exp} \left( f_a(\mathbf{x}_i^a,\mathbf{x}_j^a) \right)  }{ \sum_{j=1}^N f_s(\mathbf{x}_i^s,\mathbf{x}_j^s) ~ \mathrm{exp} \left( f_a(\mathbf{x}_i^a,\mathbf{x}_j^a) \right) } ,
\label{eq:graph_softmax}
\end{equation}
where we perform normalization on each actor node using softmax function so that the sum of all the relation values of one actor node $i$ will be $1$.

{\bf Appearance relation.}
Here we discuss different choices for computing appearance relation value between actors:

\begin{enumerate} [fullwidth, itemindent=1.2em, nosep, label=(\arabic*)]

\item {\em Dot-Product}: The dot-product similarity of appearance features can be considered as a simple form of relation value. It is computed as:
\begin{equation}
    f_a(\mathbf{x}_i^a,\mathbf{x}_j^a)=\frac{(\mathbf{x}_i^a)^\mathrm{T} \mathbf{x}_j^a}{ \sqrt{d} },
\end{equation}
where $\sqrt{d}$ acts as a normalization factor. 
    
\item {\em Embedded Dot-Product}: Inspired by the Scaled Dot-Product Attention mechanism~\cite{cite_s_1}, we can extend the dot-product operation to compute similarity in an embedding space, and the corresponding function can be expressed as:
\begin{equation}
    f_a(\mathbf{x}_i^a,\mathbf{x}_j^a)=\frac{\theta(\mathbf{x}_i^a)^\mathrm{T}\phi(\mathbf{x}_j^a)}{\sqrt{d_k}},
\label{eq:embedded_dot_product}
\end{equation}
where $\theta(\mathbf{x}_i^a)=\mathbf{W}_\theta \mathbf{x}_i^a+\mathbf{b}_\theta$ and $\phi(\mathbf{x}_j^a)=\mathbf{W}_\phi \mathbf{x}_j^a+\mathbf{b}_\phi$ are two learnable linear transformations. $\mathbf{W}_\theta \in \mathbb{R}^{d_k \times d}$ and $\mathbf{W}_\phi \in \mathbb{R}^{d_k \times d}$ are weight matrices, $\mathbf{b}_\theta \in \mathbb{R}^{d_k}$ and $\mathbf{b}_\phi \in \mathbb{R}^{d_k}$ are weight vectors. By learnable transformations of original features, we can learn the relation value between two actors in a subspace.
    
\item {\em Relation Network}: We also evaluate the Relation Network module proposed in~\cite{cite_r_16}. It can be written as:
\begin{equation}
    f_a(\mathbf{x}_i^a,\mathbf{x}_j^a)=\mathrm{ReLU} \left( \mathbf{W}[\theta(\mathbf{x}_i^a), \phi(\mathbf{x}_j^a)]+\mathbf{b} \right) ,
\end{equation}
where $[ \cdot,\cdot ]$ is the concatenation operation and $\mathbf{W}$ and $\mathbf{b}$ are learnable weights that project the concatenated vector to a scalar, followed by a ReLU non-linearity.

\end{enumerate}

{\bf Position relation.}
In order to add spatial structural information to actor graph, the position relation between actors needs to be considered. To this end, we investigate two approaches to use spatial features in our work:

\begin{enumerate} [fullwidth, itemindent=1.2em, nosep, label=(\arabic*)]

\item {\em Distance Mask}: Generally, signals from local entities are more important than the signals from distant entities. And the relation information in the local scope has more significance than global relation for modeling the group activity. Based on these observations, we can set $\mathbf{G}_{ij}$ as zero for two actors whose distance is above a certain threshold. We call the resulted ARG as {\em localized ARG}. The $f_s$ is formed as:
\begin{equation}
    f_s(\mathbf{x}_i^s,\mathbf{x}_j^s)=\mathbb{I} \left( d(\mathbf{x}_i^s,\mathbf{x}_j^s) \leq \mu  \right) ,
\end{equation}
where $\mathbb{I}(\cdot)$ is the indicator function, $d(\mathbf{x}_i^s,\mathbf{x}_j^s)$ denotes the Euclidean distance between center points of two actors' bounding boxes, and $\mu$ acts as a distance threshold which is a hyper-parameter.

\item {\em Distance Encoding}: Alternatively, we can use the recent approaches~\cite{cite_s_1} for learning position relation. Specifically, the position relation value is computed as
\begin{equation}
    f_s(\mathbf{x}_i^s,\mathbf{x}_j^s)=\mathrm{ReLU} \left( \mathbf{W}_{s}\mathcal{E}(\mathbf{x}_i^s,\mathbf{x}_j^s)+\mathbf{b}_s \right) ,
\end{equation}
the relative distance between two actors is embedded to a high-dimensional representation by $\mathcal{E}$, using cosine and sine functions of different wavelengths. The feature dimension after embedding is $d_s$. We then transform the embedded feature into a scalar by weight vectors $\mathbf{W}_{s}$ and $\mathbf{b}_s$, followed by a ReLU activation.

\end{enumerate}


{\bf Multiple graphs.}
A single ARG $\mathbf{G}$ typically focuses on a specific relation signal between actors, therefore discarding a considerable amount of context information. In order to capture diverse types of relation signals, we can extend the single actor relation graph into multiple graphs. That is, we build a group of graphs $\mathcal{G}=(\mathbf{G}^1,\mathbf{G}^2,\cdots,\mathbf{G}^{N_g})$ on a same actors set,where $N_g$ is the number of graphs. 
Every graph $\mathbf{G}^i$ is computed in the same way according to \myref{eq:graph_softmax}, but with unshared weights. Building multiple relation graphs allows the model to jointly attend to different types of relation between actors. Hence, the model can make more robust relational reasoning upon the graphs.

{\bf Temporal modeling.}
Temporal context information is a crucial cue for activity recognition. Different from prior works, which employ Recurrent Neural Network to aggregate temporal information on dense frames, our model merges the information in the temporal domain via a sparse temporal sampling strategy~\cite{cite_o_1}. During training, we randomly sample a set of $K=3$ frames from the entire video, and build temporal graphs upon the actors in these frames. We call the resulted ARG as {\em randomized ARG}. At testing time, we can use a sliding window approach, and the activity scores from all windows are mean-pooled to form global activity prediction.

Empirically we find that sparsely sampling frames when training yields significant improvements on recognition accuracy. A key reason is that, existing group activity recognition datasets (e.g., Collective Activity dataset and Volleyball dataset) remain limited, in both size and diversity. Therefore, randomly sampling the video frames results in more diversity during training and reduces the risk of over-fitting. Moreover, this sparse sampling strategy preserves temporal information with dramatically lower cost, thus enabling end-to-end learning under a reasonable budget in both time and computing resources.

\subsection{Reasoning and Training on Graphs}

Once the ARGs are built, we can perform relational reasoning on them for recognizing individual actions and group activity. We first review a graph reasoning module, called Graph Convolutional Network (GCN)~\cite{cite_g_10}. GCN takes a graph as input, performs computations over the structure, and returns a graph as output, which can be considered as a ``graph-to-graph" block. For a target node $i$ in the graph, it aggregates features from all neighbor nodes according to the edge weight between them. Formally, one layer of GCN can be written as:
\begin{equation}
    \mathbf{Z}^{(l+1)}=\sigma \left( \mathbf{G}\mathbf{Z}^{(l)}\mathbf{W}^{(l)} \right) ,
\end{equation}
where $\mathbf{G} \in \mathbb{R}^{N \times N}$ is the matrix representation of the graph. $\mathbf{Z}^{(l)} \in \mathbb{R}^{N \times d}$ is the feature representations of nodes in the $l^{th}$ layer, and $\mathbf{Z}^{(0)}=\mathbf{X}$. $\mathbf{W}^{(l)} \in \mathbb{R}^{d \times d}$ is the layer-specific learnable weight matrix. $\sigma (\cdot)$ denotes an activation function, and we adopt ReLU in this work. This layer-wise propagation can be stacked into multi-layers. For simplicity, we only use a layer of GCN in this work.

The original GCN operates on a single graph structure.
After GCN, the way to fuse a group of graphs together remains an open question. In this work, we employ the late fusion scheme, namely fuse the features of same actor in different graphs after GCN:
\begin{equation}
    \mathbf{Z}^{(l+1)} = \sum^{N_g}_{i=1}{\sigma \left( \mathbf{G}^i\mathbf{Z}^{(l)}\mathbf{W}^{(l,i)} \right)} ,
\end{equation}
where we employ element-wise sum as a fusion function. We also evaluate concatenation as fusion function. Alternatively, a group of graphs can also be fused by early fusion, that is, fused via summation to one graph before GCN. We compare different methods of fusing a group of graphs in our experiments.

Finally the output relational features from GCN are fused with original features via summation to form the scene representation. As illustrated in Figure~\ref{fig:figure_2}, the scene representation is fed to two classifiers to generate individual actions and group activity predictions.

The whole model can be trained in an end-to-end manner with backpropagation. 
Combining with standard cross-entropy loss, the final loss function is formed as
\begin{equation}
    \mathcal{L}=\mathcal{L}_1(y^{G},\hat{y}^{G}) + \lambda\mathcal{L}_2(y^{I},\hat{y}^{I}) ,
\end{equation}
where $\mathcal{L}_1$ and $\mathcal{L}_2$ are the cross-entropy loss, $y^{G}$ and $y^{I}$ denote the ground-truth labels of group activity and individual action, $\hat{y}^{G}$ and $\hat{y}^{I}$ are the predictions to group activity and individual action. The first term corresponds to group activity classification loss, and the second is the loss of the individual action classification. The weight $\lambda$ is used to balance these two tasks.

\begin{table}

\begin{subtable}{0.48\textwidth}
    \begin{center}
    \begin{tabular}{|l|c|}
    \hline
    Method & Accuracy \\
    \hline\hline
    base model & 89.8\% \\
    \hline
    dot-product & {\bf 91.3\%} \\
    embedded dot-product & {\bf 91.3\%} \\
    relation network & 90.7\% \\
    \hline
    \end{tabular}
    \caption{Exploration of different appearance relation functions.}
    \label{table:table_1a}
    \end{center}
    \label{}
\end{subtable}

\begin{subtable}{0.48\textwidth}
    \begin{center}
    \begin{tabular}{|l|c|}
    \hline
    Method & Accuracy \\
    \hline\hline
    no position relation  & 91.3\% \\
    \hline
    distance mask & {\bf 91.6\%} \\
    distance encoding & 91.5\% \\
    \hline
    \end{tabular}
    \caption{Exploration of different position relation functions.}
    \label{table:table_1b}
    \end{center}
    \label{}
\end{subtable}


\begin{subtable}{0.48\textwidth}
    \begin{center}
    \resizebox{\linewidth}{!}{
    \begin{tabular}{|l|c|c|c|c|c|}
    \hline
    Number & 1 & 4 & 8 & 16 & 32   \\
    \hline
    Accuracy & 91.6\% & 92.0\% & 92.0\% & {\bf 92.1\%} & 92.0\%   \\
    \hline
    \end{tabular}
    }
    \caption{Exploration of number of graphs.}
    \label{table:table_1c}
    \end{center}
    \label{}
\end{subtable}

\begin{subtable}{0.48\textwidth}
    \begin{center}
    \begin{tabular}{|l|c|}
    \hline
    Method & Accuracy \\
    \hline\hline
    early fusion & 90.8\% \\
    late fusion (summation) & {\bf 92.1\%} \\
    late fusion (concatenation) & 91.9\% \\
    \hline
    \end{tabular}
    \caption{Exploration of different methods for fusing multiple graphs.}
    \label{table:table_1d}
    \end{center}
    \label{}
\end{subtable}

\begin{subtable}{0.48\textwidth}
    \begin{center}
    \begin{tabular}{|l|c|}
    \hline
    Method & Accuracy \\
    \hline\hline
    single frame & 92.1\% \\
    \hline
    TSN (3 frames) & 92.3\% \\
    temporal-graphs (3 frames) & {\bf 92.5\%} \\
    \hline
    \end{tabular}
    \caption{Exploration of temporal modeling methods.}
    \label{table:table_1e}
    \end{center}
\end{subtable}

\caption{Ablation studies for group activity recognition accuracy on the Volleyball dataset.}
\label{table:table_1}
\end{table}

\section{Experiments}
\label{section:Experiments}
In this section, we first introduce two widely-adopted datasets and the implementation details of our approach. Then, we perform a number of ablation studies to understand the effects of proposed components in our model. We also compare the performance of our model with the state of the art methods. Finally, we visualize our learned actor relation graphs and features.

\subsection{Datasets and Implementation Details}
{\bf Datasets.}
We conduct experiments on two publicly available group activity recognition datasets, namely the Volleyball dataset and the Collective Activity dataset. 

The Volleyball dataset~\cite{cite_d_2} is composed of 4830 clips gathered from 55 volleyball games, with 3493 training clips and 1337 for testing. Each clip is labeled with one of 8 group activity labels (right set, right spike, right pass, right winpoint, left set, left spike, left pass and left winpoint). Only the middle frame of each clip is annotated with the players' bounding boxes and their individual actions from 9 personal action labels (waiting, setting, digging, failing, spiking, blocking, jumping, moving and standing). Following~\cite{cite_a_2}, we use 10 frames to train and test our model, which corresponds to 5 frames before the annotated frame and 4 frames after. To get the ground truth bounding boxes of unannotated frames, we use the tracklet data provided by~\cite{cite_a_1}.

The Collective Activity dataset~\cite{cite_d_1} contains 44 short video sequences (about 2500 frames) from 5 group activities (crossing, waiting, queueing, walking and talking) and 6 individual actions (NA, crossing, waiting, queueing, walking and talking). The group activity label for a frame is defined by the activity in which most people participate. We follow the same evaluation scheme of~\cite{cite_a_18} and select $1/3$ of the video sequences for testing and the rest for training.

{\bf Implementation details.}
We extract 1024-dimensional feature vector for each actor with ground-truth bounding boxes, using the methods mentioned in Section~\ref{section:Framework}.
During ablation studies, we adopt Inception-v3 as backbone network. We also experiment with VGG~\cite{cite_o_6} network for fair comparison with prior methods.
Due to memory limits, we train our model in two stages: first, we fine-tune the ImageNet pre-trained model on single frame randomly selected from each video without using GCN.
We refer to the fine-tuned model described above as our base model throughout experiments.
The base model performs group activity and individual action classification on original features of actors without relational reasoning.
Then we fix weights of the feature extraction part of network, and further train the network with GCN. 

We adopt stochastic gradient descent with ADAM to learn the network parameters with fixed hyper-parameters to $\beta_1=0.9,\beta_2=0.999,\epsilon=10^{-8}$. 
For the Volleyball dataset, we train the network in 150 epochs using mini-batch size of 32 and a learning rate ranging from $0.0002$ to $0.00001$.
For the Collective Activity dataset, we use mini-batch size of 16 with a learning rate of $0.0001$, and train the network in 80 epochs.
The individual action loss weight $\lambda=1$ is used.
Besides, the parameters of the GCN are set as $d_k=256,d_s=32$, and we adopt the $1/5$ of the image width to be the distance mask threshold $\mu$. 

Our implementation is based on PyTorch deep learning framework.
The running time for inferring a video is approximately 0.2s on a single TITAN-XP GPU.

\subsection{Ablation Studies}
In this subsection, we perform detailed ablation studies on the Volleyball dataset to understand the contributions of the proposed model components to relation modeling using group activity recognition accuracy as evaluation metric. The results are shown in Table~\ref{table:table_1}.

{\bf Appearance relation.} 
We begin our experiments by studying the effect of modeling the appearance relation between actors and different functions to compute appearance relation value. Based on single frame, we build single ARG without using position relation. The results are listed in Table~\ref{table:table_1a}. 
We first observe that explicitly modeling the relation between actors brings significant performance improvement. All models with GCN outperform the base model. 
Then it is shown that the dot-product and embedded dot-product yield same recognition accuracy of $91.3\%$, and perform better than the relation network.
We conjecture that dot-product operation is more stable for representing relation information.
In the following experiments, embedded dot-product is used to compute appearance relation value.

{\bf Position relation.}
We further add spatial structural information to ARG. In Section~\ref{section:Graph}, we present two methods to use spatial features: distance mask and distance encoding.
Results on comparing the performance of these two methods are reported in Table~\ref{table:table_1b}. 
We can see that these two methods both obtain better performance than those without using spatial features, demonstrating the effectiveness of modeling position relation.
And the distance mask yields slightly better accuracy than distance encoding.
In the rest of the paper, we choose distance mask to represent position relation.

{\bf Multiple graphs.}
We also investigate the effectiveness of building a group of graphs to capture different kinds of relation information. 
First, we compare the performance of using different number of graphs.
As shown in Table~\ref{table:table_1c}, we observe that building multiple graphs leads to consistent and significant gain compared with only building single graph, and is able to further boost accuracy from $91.6\%$ to $92.1\%$. 
Then we evaluate three methods to fuse a group of graphs: (1) early fusion, (2) late fusion via summation, (3) late fusion via concatenation. The results of experiments using 16 graphs are summarized in Table~\ref{table:table_1d}. We see that the late fusion via summation achieves the best performance.
We note that the early fusion scheme, which aggregates a group of graphs by summation before GCN, results in the performance drops dramatically.
This observation indicates that the relation values learned by different graphs encode different semantic information and will cause confusion for relational reasoning if they are fused before graph convolution.
We adopt $N_g=16$ and late fusion via summation in the following experiments.

{\bf Temporal modeling.}
With all the design choices set, we now extend our model to temporal domain. As mentioned in Section~\ref{section:Graph}, we employ sparse temporal sampling strategy~\cite{cite_o_1}, and uniformly sample a set of $K=3$ frames from the entire video during training.
In the simplest setting, we can handle the input frames separately, then fuse the prediction scores of different frames as Temporal Segment Network (TSN)~\cite{cite_o_1}. 
Alternatively, we can build temporal graphs upon the actors in input frames and fuse temporal information by GCN.
We report the accuracies of these two temporal modeling methods in Table~\ref{table:table_1e}.
We see that TSN modeling is helpful to improve the performance of our model.
Moreover, building temporal graphs further boosts accuracy to $92.5\%$, which demonstrates that temporal reasoning helps to differentiate between group activity categories.

\begin{table}[t]
\begin{subtable}{0.48\textwidth}
\begin{center}
\resizebox{\linewidth}{!}{
\begin{tabular}{|l|c|c|c|}
\hline

\multirow{2}*{Method} & \multirow{2}*{Backbone} & Group & Individual \\
		~ & ~ & activity & action \\


\hline\hline
HDTM~\cite{cite_a_2}     & AlexNet & 81.9\% & - \\
CERN~\cite{cite_a_4}     & VGG16   & 83.3\% & - \\
stagNet (GT)~\cite{cite_a_18} & VGG16   & 89.3\% & - \\
stagNet (PRO)~\cite{cite_a_18} & VGG16   & 87.6\% & - \\
HRN~\cite{cite_a_17}     & VGG19   & 89.5\% & - \\
SSU (GT)~\cite{cite_a_1} & Inception-v3  & 90.6\% & 81.8\% \\
SSU (PRO)~\cite{cite_a_1} & Inception-v3 & 86.2\% & 77.4\% \\
\hline\hline

OURS (GT) & Inception-v3   & 92.5\% & 83.0\% \\
OURS (PRO) & Inception-v3   & 91.5\% & - \\
OURS (GT) & VGG16   & 91.9\% & \bf{83.1}\% \\
OURS (GT) & VGG19   & \bf{92.6\%} & 82.6\% \\
\hline

\end{tabular}
}
\caption{Comparison with state of the art on the Volleyball dataset.}
\label{table:table_2a}
\end{center}
\label{}
\end{subtable}

\begin{subtable}{0.48\textwidth}
\begin{center}
\resizebox{\linewidth}{!}{
\begin{tabular}{|l|c|c|}
\hline
Method & Backbone & Group activity  \\
\hline\hline

SIM~\cite{cite_a_5}      & AlexNet & 81.2\%  \\
HDTM~\cite{cite_a_2}     & AlexNet & 81.5\%  \\
Cardinality Kernel~\cite{cite_a_19}     & None & 83.4\%  \\
SBGAR~\cite{cite_a_6}     & Inception-v3 & 86.1\%  \\
CERN~\cite{cite_a_4}         & VGG16 & 87.2\%  \\
stagNet (GT)~\cite{cite_a_18}     & VGG16 & 89.1\%  \\
stagNet (PRO)~\cite{cite_a_18}     & VGG16 & 87.9\%  \\

\hline\hline

OURS (GT) & Inception-v3 &  \bf{91.0\%} \\
OURS (PRO) & Inception-v3 &  90.2\% \\
OURS (GT) & VGG16        &  90.1\% \\
\hline

\end{tabular}
}
\end{center}
\caption{Comparison with state of the art on the Collective dataset.}
\label{table:table_2b}
\end{subtable}
\label{}
\vspace{-2mm}
\caption{Comparison with state of the art methods. GT and PRO indicate using ground-truth and proposal-based bounding boxes, respectively.}
\vspace{-2mm}
\label{table:table_2}
\end{table}

\begin{figure*}[t]

\setlength{\abovecaptionskip}{-0.3cm}
\setlength{\belowcaptionskip}{-0.2cm}

\begin{center}
\includegraphics[width=0.92\linewidth]{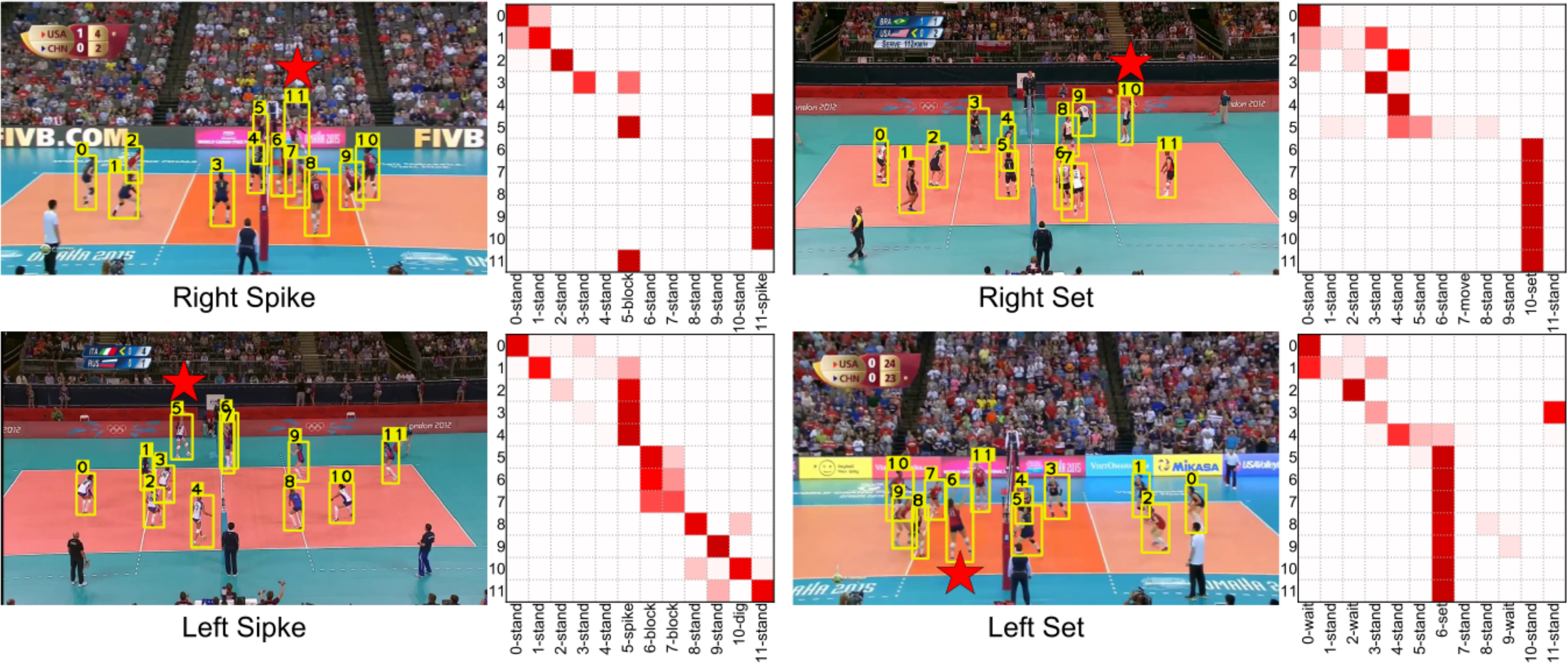}
\end{center}
\caption{Visualization of learned actor relation graphs. 
Each row shows two examples.
For each example, we plot: (1) input frame with group-truth bounding boxes and group activity label;  (2) matrix $\mathbf{G}$ of learned relation graph with ground-truth individual action labels. The actor who has max column sum of $\mathbf{G}$ in each frame is denoted with red star.
}
\label{fig:visual}
\end{figure*}

\begin{figure*}

\setlength{\abovecaptionskip}{-0.3cm}
\setlength{\belowcaptionskip}{-0.2cm}

\begin{center}
\includegraphics[width=0.92\linewidth]{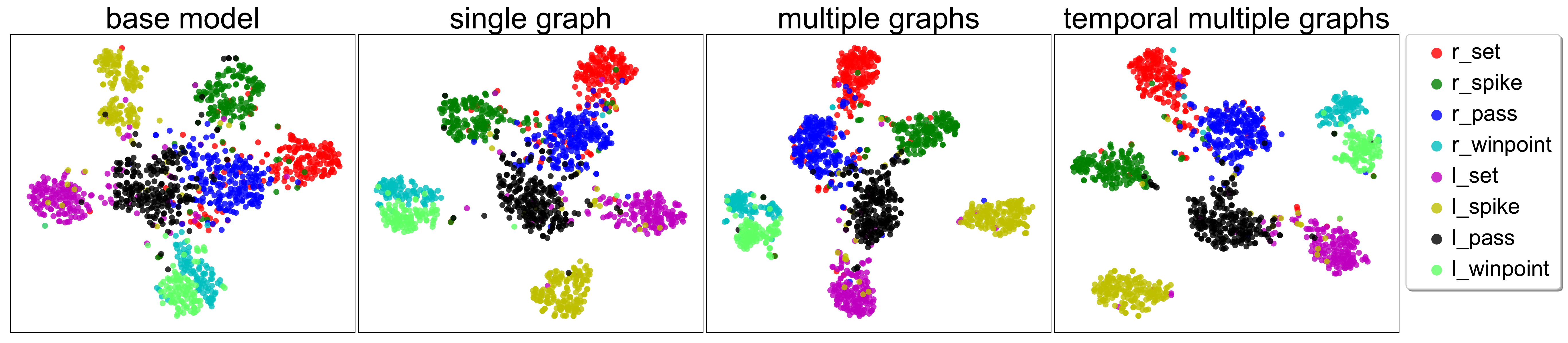}
\end{center}
\caption{t-SNE~\cite{cite_o_2} visualization of embedding of video representation on the Volleyball dataset learned by different model variants: base model, single graph, multiple graphs, temporal multiple graphs. Each video is visualized as one point and colors denote different group activities (better view in color version).}
\label{fig:tsne}
\end{figure*}

\subsection{Comparison with the State of the Art}
Now, we compare our best models with the state-of-the-art methods in Table~\ref{table:table_2}.
For fair comparison with prior methods, we report our results with both Inception-v3 and VGG backbone network.
Meanwhile, we perform proposal-based experiment.
We train a Faster-RCNN~\cite{cite_o_7} with training data.
Using the bounding boxes from Faster-RCNN at testing time, our model can still achieve promising accuracy.

Table~\ref{table:table_2a} shows the comparison with previous results on the Volleyball dataset for group activity and individual action recognition. 
Our method surpasses all the existing methods by a good margin, establishing the new state-of-the-art.
Our model with Inception-v3 utilizes the same feature extraction strategy as~\cite{cite_a_1}, and outperforms it by about $2\%$ on group activity recognition accuracy, since our model can capture and exploit the relation information among actors. 
And, we also achieve better performance on individual action recognition task.
Meanwhile, our method outperforms the recent methods using hierarchical relational networks~\cite{cite_a_17} or semantic RNN~\cite{cite_a_18}, mostly because we explicitly model the appearance and position relation graph, and adopt more efficient temporal modeling method.

We further evaluate the proposed model on the Collective Activity dataset. The results and comparison with previous methods are listed in Table~\ref{table:table_2b}. Our temporal multiple graphs model again achieves the state-of-the-art performance with $91.0\%$ group activity recognition accuracy.
This outstanding performance shows the effectiveness and generality of proposed ARG for capturing the relation information in multiple people scene.

\subsection{Model Visualization}
{\bf Actor relation graph visualization}
We visualize several examples of the relation graph generated by our model in Figure~\ref{fig:visual}.
We use the single graph model on single frame, because it is easier to visualize.
Visualization results facilitate us understanding how ARG works.
We can see that our model is able to capture relation information for group activity recognition, and the generated ARG can automatically discover the key actor to determine the group activity in the scene.

{\bf t-SNE visualization of the learned representation.}
Figure~\ref{fig:tsne} shows the t-SNE~\cite{cite_o_2} visualization for embedding the video representation learned by different model variants. Specifically, we project the representations of videos on the validation set of Volleyball dataset into 2-dimensional space using t-SNE. 
We can observe that the scene-level representations learned by using ARG are better separated.
Moreover, building multiple graphs and aggregating temporal information lead to better differentiate group activities.
These visualization results indicate our ARG models are more effective for group activity recognition.

\vspace{-3mm}
\section{Conclusion}
This paper has presented a flexible and efficient approach to determine relevant relation between actors in a multi-person scene.
We learn {\em Actor Relation Graph} (ARG) to perform relational reasoning on graphs for group activity recognition.
We also evaluate the proposed model on two datasets and establish new state-of-the-art results.
The comprehensive ablation experiments and visualization results show that our model is able to learn relation information for understanding group activity. 
In the future, we plan to further understand how ARG works, and incorporate more global scene information for group activity recognition.

\vspace{-2mm}
\section*{Acknowledgement}
This work is supported by the National Science Foundation of China under Grant No.61321491, and Collaborative Innovation Center of Novel Software Technology and Industrialization.

{\small
\bibliographystyle{ieee}
\bibliography{mybib}
}

\end{document}